\documentclass[10pt,twocolumn,letterpaper]{article}

\usepackage[pagenumbers]{cvpr} 

\usepackage[dvipsnames]{xcolor}

\usepackage{bm}
\usepackage{multirow}
\usepackage{graphicx}
\usepackage{colortbl}
\definecolor{lightgray}{gray}{0.9}
\usepackage{float} 

\usepackage[acronym]{glossaries}
\newacronym{fps}{fps}{frames per second}
\newacronym{cdm}{CDM}{Cascaded Diffusion Model}
\newacronym{edm}{EDM}{Elucidated Diffusion Model}
\newacronym{ml}{ML}{Machine Learning}
\newacronym{vae}{VAE}{Variational Auto-Encoder}
\newacronym[longplural={Denoising Diffusion Probabilistic Models},plural={DDPMs}]{ddpm}{DDPM}{Denoising Diffusion Probabilistic Model}
\newacronym[longplural={Denoising Diffusion Implicit Models},plural={DDIMs}]{ddim}{DDIM}{Denoising Diffusion Implicit Model}
\newacronym{adm}{ADM}{Ablated Diffusion Model}
\newacronym{pndm}{PNDM}{Pseudo Numerical methods for diffusion models}
\newacronym[longplural={Generative Adversarial Networks},plural={GANs}]{gan}{GAN}{Generative Adversarial Network}
\newacronym{vqvae}{VQ-VAE}{Vector Quantized Variational Autoencoder}
\newacronym{lidm}{LIDM}{Latent Image Diffusion Model}
\newacronym{lvdm}{LVDM}{Latent Video Diffusion Model}
\newacronym[longplural={Stochastic Differential Equations},plural={SDEs}]{sde}{SDE}{Stochastic Differential Equation}
\newacronym[longplural={Ordinary Differential Equations},plural={ODEs}]{ode}{ODE}{Ordinary Differential Equation}
\newacronym{vdm}{VDM}{Video Diffusion Model}
\newacronym{fid}{FID}{Fréchet Inception Distance}
\newacronym{is}{IS}{Inception Score}
\newacronym{fvd}{FVD}{Fréchet Video Distance}

\definecolor{cvprblue}{rgb}{0.21,0.49,0.74}
\usepackage[pagebackref,breaklinks,colorlinks,citecolor=cvprblue]{hyperref}

\def\paperID{3304} 
\def\confName{CVPR}
\def\confYear{2024}

\title{JVID: Joint Video-Image Diffusion for Visual-Quality and Temporal-Consistency in Video Generation}

\author{Hadrien Reynaud\\
Imperial College London\\
London, UK\\
{\tt\small hjr119@imperial.ac.uk}
\and
Matthew Baugh\\
Imperial College London\\
London, UK\\
\and
Mischa Dombrowski\\
Friedrich-Alexander-Universität\\
Erlangen-Nürnberg, DE\\
\and
Sarah Cechnicka\\
Imperial College London\\
London, UK\\
\and
Qingjie Meng\\
University of Birmingham\\
Birmingham UK\\
\and
Bernhard Kainz\\
Imperial College London\\
London, UK\\
Friedrich-Alexander-Universität\\
Erlangen-Nürnberg, DE\\
}

\begin{document}
\maketitle
\begin{abstract}
We introduce the Joint Video-Image Diffusion model (JVID), a novel approach to generating high-quality and temporally coherent videos. We achieve this by integrating two diffusion models: a Latent Image Diffusion Model (LIDM) trained on images and a Latent Video Diffusion Model (LVDM) trained on video data. Our method combines these models in the reverse diffusion process, where the LIDM enhances image quality and the LVDM ensures temporal consistency. This unique combination allows us to effectively handle the complex spatio-temporal dynamics in video generation. Our results demonstrate quantitative and qualitative improvements in producing realistic and coherent videos.
\end{abstract}    
\section{Introduction}
\label{sec:intro}

Large-scale diffusion-based text-to-image models have led to remarkable improvements in image generation~\cite{sohl_dickstein2015deep,song2019generative,ho2020denoising,ramesh2022hierarchical,saharia2022photorealistic,balaji2022ediff,rombach2022high,gu2022vector}.
They can produce high-quality, photorealistic images that follow complex text descriptions with great accuracy.
Due to the success of these text-to-image generations, several works have experimented with applying them to video generation.
However, generating videos is much more challenging because of their high-dimensionality and complex spatio-temporal dynamics.

Many previous works explored various generative models for video generation, such as \glspl{gan}~\cite{skorokhodov2021stylegan,tian2021good,saito2017temporal,singer2022make,vondrick2016generating,yu2022generating} and autoregressive models~\cite{yan2021videogpt,esser2021taming,rakhimov2020latent,vaswani2017attention,weissenborn2019scaling}. Nonetheless, these methods have difficulties achieving wide video mode coverage, maintaining long-term dependency, and high visual quality.
Most concurrent works use diffusion models for video generation~\cite{ho2022video,voleti2022masked,harvey2022flexible,zhou2022magicvideo,wu2022tune,blattmann2023align,khachatryan2023text2video,hoeppe2022diffusion,yang2022diffusion,nikankin2022sinfusion,luo2023videofusion,an2023latent,wang2023videofactory,singer2022make,ho2022imagen,zhou2022magicvideo}.
These approaches achieve better video quality than \gls{gan}-based methods, but they come with a substantial computational cost, long training periods and long iterative sampling.

\begin{figure}
    \centering
    \includegraphics[width=\linewidth]{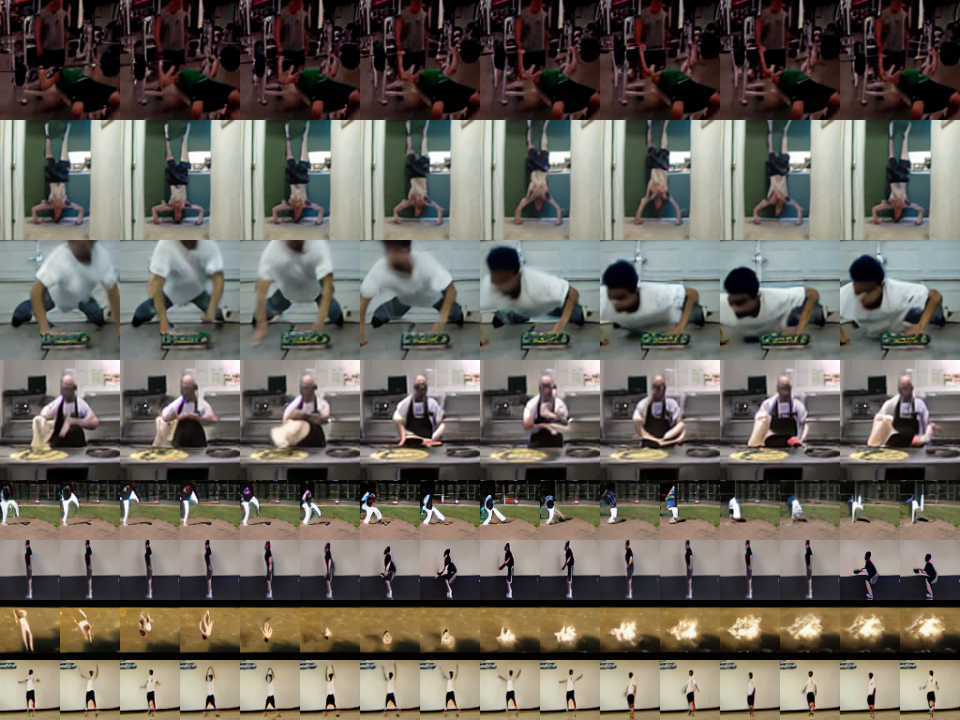}
    \caption{Video samples generated with our JVID model, combining both an image and a video diffusion model during sampling, to produce high quality and temporally coherent videos. Rows 1-4 are generated at $128 \times 128$, while rows 5-8 are $64 \times 64$.}
    \label{fig:samples}
\end{figure}

In this work, we introduce a novel \textbf{J}oint \textbf{V}ideo-\textbf{I}mage \textbf{D}iffusion model (JVID), to generate realistic-looking and temporally-coherent videos. Examples from JVID are shown in \Cref{fig:samples}. 
We present a new sampling strategy where we generate videos by mixing a video diffusion model and an image diffusion model.
More specifically, two diffusion models, \emph{i.e.,} an image model and a video model, are trained independently under the same diffusion framework. 
We then derive a method to sample from both models during the iterative reverse process, in order to generate videos with better visual quality.
The idea is that at every denoising step, we denoise our sample with one model or the other. 
Because the models are trained within the same framework, they are compatible with each other, and therefore, the underlying iterative denoising process is coherent.

Prior works either train diffusion-based video models by adapting a pre-trained image diffusion model~\cite{blattmann2023align,wu2022tune} or train video diffusion models from scratch~\cite{ho2022video,singer2022make,he2022latent}. Our work utilises both an image and a video model, but their individual training regime does not matter, as long as they are both trained using the same perturbation process and noise schedule, as we detail in \Cref{sec:methods}.
Therefore, we do not require any special tricks during training, which means that any pair of pre-trained models, respecting the two previously stated conditions, would be compatible with our proposed sampling method.

Furthermore, inspired by latent diffusion models~\cite{rombach2022high,gu2022vector}, our method models image and video distributions in a latent space, which greatly reduces the computational cost of training and the sampling time.

The contributions of our work can be summarized as:
\begin{enumerate}
    \item We propose a new sampling strategy to leverage the benefits of multiple diffusion models during inference.
    \item We introduce two techniques (inference entropy reduction and temporal latent smoothing) to further enhance the temporal consistency in video diffusion models and,
    \item we release pre-trained models to enable future research in the domain of video-image generation.
\end{enumerate}

\section{Related Works}
\label{sec:rel_works}

\subsection{Diffusion models}

A diffusion model is a type of score-based generative model which learns a mapping from a Gaussian noise distribution to a target data distribution~\cite{sohl_dickstein2015deep,song2019generative,ho2020denoising}.
With stable training and great scalability, diffusion models can achieve both high sample quality and great distribution coverage~\cite{dhariwal2021diffusion,saharia2022photorealistic,ramesh2021zero,rombach2022high}, leading to remarkable progress in text-to-image generation tasks. 
A seminal contribution in the field of diffusion models is the \gls{ddpm} as presented in Ho et al.~\cite{ho2020denoising}.
This model capitalizes on the connection between Langevin dynamics~\cite{sohl_dickstein2015deep} and denoising score matching~\cite{song2020score}, employing them to construct a weighted variational bound that serves as the basis for the optimization process.
Based on the \gls{ddpm}, other methods were proposed to improve the generation process. 
For example, the \gls{ddim}~\cite{song2020denoising} accelerates the sampling process while maintaining the \gls{ddpm} training process. The \gls{adm}~\cite{dhariwal2021diffusion} uses an improved architecture and classifier guidance to achieve increased image quality. 
\gls{pndm}~\cite{liu2022pseudo} establishes a theoretical connection between DDPM and numerical methods.
By changing the classical numerical methods to pseudo numerical methods, \gls{pndm} can accelerate the inference process while maintaining the quality of the generated images.
In this work, we use \gls{ddpm} sampling to benefit from the sample variety allowed by the stochastic sampling, and we rely on \gls{pndm} to train our models.

\subsection{Diffusion-based Image Models}

Diffusion models facilitate large-scale text-to-image generation~\cite{ramesh2022hierarchical,saharia2022photorealistic,nichol2021glide}. 
However, they still suffer from major limitations \emph{i.e.}, high memory consumption, high computational complexity and long sampling time, which is orders of magnitude longer than other generative methods due to their iterative nature. 
Some approaches significantly lessened these issues by applying the denoising process in the latent space rather than the RGB-space.
For example, \cite{rombach2022high} trains a diffusion model in the latent space of a powerful pre-trained autoencoder, while \cite{gu2022vector} uses a pre-trained \gls{vqvae}.
These methods drastically reduce the memory footprint of diffusion models, resulting in faster training and inference, while keeping the quality and mode coverage of RGB-space image diffusion models. 
In particular, \cite{rombach2022high} opened the path for many subsequent works, thanks to its large-scale implementation known as Stable Diffusion\footnote{https://stability.ai/stable-diffusion}.
Many following works use the pre-trained Stable Diffusion weights and fine-tune them on specific tasks \cite{ruiz2022dreambooth,zhang2023adding,brooks2023instructpix2pix,kumari2022multi,wu2022tune,ma2023follow}, while others focused on adding more control to the generation process, given that text can be limiting \cite{meng2022sdedit,hertz2022prompt,parmar2023zero,qi2023fatezero,ge2023expressive,ceylan2023pix2video,bar_tal2022text2live,molad2023video}. 
In this work, we train latent image diffusion models, as they are the most direct way to obtain the sample quality that we aim for, at an adequate computational cost.
We rely on their reduced memory requirements to train models at different resolutions.

\subsection{Diffusion-based Video Models}
Many existing works have attempted to extend image \glspl{gan} to generate videos~\cite{skorokhodov2021stylegan,tian2021good,saito2017temporal,singer2022make,vondrick2016generating,yu2022generating}. 
However, \glspl{gan} often suffer from mode collapse and unstable training. 
Some other approaches use autoregressive models for video generation~\cite{yan2021videogpt,esser2021taming,rakhimov2020latent,vaswani2017attention,weissenborn2019scaling}. 
Although they have outperformed pure \gls{gan}-based approaches, they tend to accumulate errors over time, resulting in poor long-term dependency. 
Recently, several works started studying diffusion models for video generation \cite{ho2022video,voleti2022masked,harvey2022flexible,zhou2022magicvideo,wu2022tune,blattmann2023align,khachatryan2023text2video,hoeppe2022diffusion,yang2022diffusion,nikankin2022sinfusion,luo2023videofusion,an2023latent,wang2023videofactory,singer2022make,ho2022imagen}. 
\gls{vdm}~\cite{ho2022video} was the first to extend an image diffusion architecture to the video domain and train jointly on both image and video data for video generation. 
Make-A-Video~\cite{singer2022make} leverages joint text-image priors and develops a large-scale video generation model based on text-to-image diffusion models. 
Imagen Video~\cite{ho2022imagen} builds a cascade of video diffusion models for high-resolution video generation, by using spatial and temporal super-resolution models and introduced the v-parameterization objective for diffusion models. 
These diffusion-based video generation methods can achieve state-of-the-art results, although they also suffer from significant computation and memory requirements. 
Concurrent works explore latent diffusion models for video generation. 
MagicVideo~\cite{zhou2022magicvideo} models video distributions in a low-dimensional latent space and proposes an elaborate diffusion architecture to generate temporally coherent videos. 
Another work, Projected Latent Video Diffusion Models (PVDM)~\cite{yu2023video}, proposes a latent video generation framework with a different video factorization technique.
Specifically, it projects videos into three 2D image-like latent vectors and then uses a 2D diffusion network to model video distribution from these 2D vectors.
We build on top of recent latent video diffusion models \cite{blattmann2023align,he2022latent} and show a novel way of improving their generation, combining them with an image diffusion model during the reverse diffusion process in the latent space.



\section{Methods}
\label{sec:methods}

\begin{figure*}[h]
    \centering
    \includegraphics[width=\textwidth]{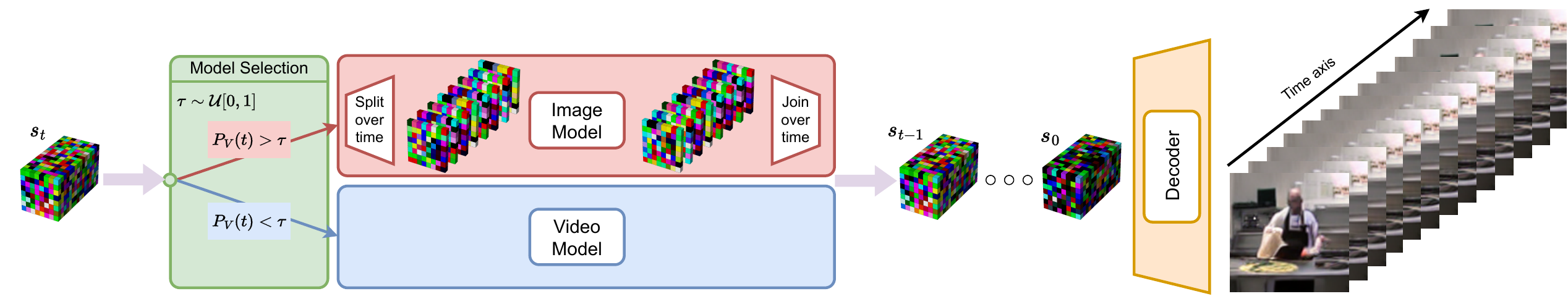}
    \caption{Our proposed mixture of denoising model sampling approach. At each step, we select one model, which is used to denoise our noise sample $\bm{s}_t$. When the sample is fully denoised, it is decoded with the VAE decoder to reconstruct the generated video frames.}
    \label{fig:method}
\end{figure*}

We combine two diffusion models during the reverse diffusion process to enforce both temporal consistency and image quality. One model is a \gls{lvdm}, trained exclusively on video data. The second model is a \gls{lidm}, trained exclusively on image data.
We summarize our sampling pipeline in \Cref{fig:method}.
Formally, we want to sample a Gaussian noise tensor $\bm{s}_{t=T} \sim \mathcal{N}(\bm{0}, \bm{I})$ of shape $F \times C \times H \times W$ and denoise it into a latent video $\bm{v} = \bm{s}_{t=0}$ of the same dimensions.
We apply a reverse diffusion process on $\bm{s}_{t=T}$, to decrease the noise level $t$ from $T$ to $0$ in an arbitrary number of steps.
During the iterative reverse diffusion process, we use either the \gls{lvdm} or the \gls{lidm} to predict the noise in $\bm{s}_{t}$, and subtract it, following the $\epsilon$-prediction from \cite{ho2020denoising}.
The \gls{lvdm} enforces temporal consistency over the time axis $F$, while the \gls{lidm} enforces better image quality over the spatial dimensions $C \times H \times W$.
In this section, we briefly present diffusion models in \Cref{sec:Diffusion}, our \gls{lidm} in \Cref{sect:lidm}, and our \gls{lvdm} in \Cref{sect:lidm}.
We then introduce our proposed mixture of denoising model sampling approach in \Cref{sect:mix}, before detailing post-processing steps to enhance the quality of our generated videos in \Cref{sec:entropy} and \Cref{sec:smoothing}.

\subsection{Denoising Diffusion Probabilistic Models}
\label{sec:Diffusion}

Both our latent image and latent video diffusion models are trained with a \gls{pndm}~\cite{liu2022pseudo} scheduler, but we use \gls{ddpm}~\cite{ho2020denoising} during inference. Here, we give an overview of the \gls{ddpm}, as the inference process is where we propose our main contribution.

Given a real sample $\bm{s}_0 \sim q(\bm{s}_0)$, the forward \gls{ddpm} process generates a Markov chain $s_1, \ldots, s_T$ by progressively injecting Gaussian noise into $\bm{s}_0$, with predetermined variance $\beta_1, \ldots, \beta_T$, \emph{i.e.},
\begin{equation}
q(\bm{s}_t|\bm{s}_{t-1}) = \mathcal{N}\left(\bm{s}_t; \sqrt{1 - \beta_t}\bm{s}_{t-1}, \beta_t \bm{I}\right).
\label{eq:ddpm1}
\end{equation}
As $t$ grows, $\bm{s}_t$ steers towards a Gaussian distribution $\mathcal{N}(\bm{0},\bm{I})$. Thus, the true posterior distribution of $q(\bm{s}_{t-1}|\bm{s}_t)$ can be approximated by $p_\theta(\bm{s}_{t-1}|\bm{s}_t)$ such that 
\begin{equation}
p_\theta(\bm{s}_{t-1}|\bm{s}_t) = \mathcal{N}\left(\bm{s}_{t-1}; \mu_\theta(\bm{s}_t), \sigma_t^2 \bm{I}\right),
\label{eq:ddpm2}
\end{equation}
where $\sigma_t$ is a known constant. 
In the reverse \gls{ddpm}, samples $\bm{s}_0~\sim~p_\theta(\bm{s}_0)$ are produced by sampling $\bm{s}_{T}~\sim~\mathcal{N}(\bm{0},\bm{I})$, and progressively reducing the noise by following the inverted Markov chain from the forward \gls{ddpm}. To do so, $p_\theta(\bm{s}_{t-1}|\bm{s}_t)$ is learned. Gaussian noise $\epsilon$ is added to $\bm{s}_0$, creating samples $\bm{s}_t \sim q(\bm{s}_t|\bm{s}_0)$, which are used to train a model $\epsilon_{\theta}$, which learns to predict $\epsilon$ through a mean-squared error loss 
\begin{equation}
\mathcal{L} = \mathbb{E}_{t \in \{1, \ldots, T\},\bm{s}_0 \sim q(\bm{s}_0), \epsilon \sim \mathcal{N}(0,\bm{I})}\left[\|\epsilon - \epsilon_\theta(\bm{s}_t, t)\|^2\right].
\label{eq:ddpm3}
\end{equation}

We derive $\mu_{\theta}(s_t)$ in \cref{eq:ddpm2} from $\epsilon_{\theta}(\bm{s}_t, t)$ to model \(p_{\theta}(\bm{s}_{t-1}|\bm{s}_t)\) \cite{ho2020denoising}. 
Following \cite{ho2022video}, the denoising model \(\epsilon_{\theta}\) is implemented via a time-conditioned UNet \cite{ronneberger2015unet} with residual layers \cite{he2016deep} and self-attention layers \cite{vaswani2017attention}. Timesteps \(t\) are passed to \(\epsilon_{\theta}\) through a sinusoidal position embedding layer \cite{vaswani2017attention}. For text conditioning, a CLIP~\cite{radford2021learning} text encoder and cross-attention layers are added, such that \(\epsilon_{\theta}(s_t, t, y)\) can be learned with no further modification~\cite{nichol2021glide,rombach2022high}. We further exploit \textit{classifier-free guidance}~\cite{ho2022classifier} for enhanced conditional generation. 
During training, the condition \(y\) in \(\epsilon_{\theta}(\bm{s}_t, t, y)\) is substituted by a null label $\emptyset$ with a fixed probability. During sampling, the model output is   
\begin{equation}
    \hat{\epsilon}_{\theta}(\bm{s}_t, t, y) = \epsilon_{\theta}(\bm{s}_t, t, \emptyset) + g \cdot (\epsilon_{\theta}(\bm{s}_t, t, y) - \epsilon_{\theta}(s_t, t, \emptyset)),
\end{equation}
where $g$ denotes the guidance scale.

Generating a new sample $\bm{s}_0$ means starting from a noise sample ${\bm{s}_{T} \sim \mathcal{N}(\bm{0}, \bm{I}})$ and progressively denoising it, by repeatedly calling $\epsilon_{\theta}(\bm{s}_t, t)$ with 
\begin{equation}
    \bm{s}_{t-1} = \frac{1}{\sqrt{\alpha_{t}}} \left( \bm{s}_{t} - \frac{\beta_t}{\sqrt{1-\Bar{\alpha}_{t}}}\epsilon_{\theta}(\bm{s}_t, t) \right) + \sigma_t\bm{z},
    \label{eq:ddpm_sampling}
\end{equation}
where $\bm{z} \sim \mathcal{N}(\bm{0}, \bm{I})$, until $t=0$.

\subsection{Latent Image Diffusion Models}
\label{sect:lidm}

For our text-to-image model, we take inspiration from the Stable Diffusion v1.5 model~\cite{rombach2022high}. Stable Diffusion is open-source, lightweight and produces state-of-the-art image quality for many scenes and styles, while operating in a latent space.
This is made possible by a pre-trained \gls{vae}, trained over billions of images \cite{schuhmann2022laion}. The \gls{vae} encodes images of dimension $3 \times H \times W$ into latent tensors of dimension $ 4 \times H/8 \times W/8$. During training, the dataset samples are encoded into that latent space, and the model learns to denoise the noisy samples in the latent space.
During inference, we start from latent Gaussian noise $\bm{s}_T \sim N(\bm{0}, \bm{I})^{4 \times H/8 \times W/8}$ and denoise it in an arbitrary number of steps, until we reach $\bm{s}_0$. $\bm{s}_0$ is then decoded using the \gls{vae} to retrieve the generated image in pixel space, as shown in \Cref{fig:method}.

For our use case, because we use a different image resolution than Stable Diffusion, we train the diffusion model from scratch, in the latent space defined by the pre-trained \gls{vae}, as well as the existing CLIP text-encoder. This also allows us to make sure that both diffusion models follow the same perturbation process and noise schedule, as required by our method.

\subsection{Latent Video Diffusion Models}
\label{sect:lvdm}

Similarly to our \gls{lidm}s, our \gls{lvdm}s are latent generative models. As such, they benefit from lower computational cost and shorter inference time. This is especially desirable for video models, where the data dimension becomes problematically large in pixel space. 

Our video model differs from the image model in terms of input/output dimensions and architecture. 
We start from the same architecture as for the \gls{lidm}, but all 2D convolutions are replaced with 3D spatial kernels $k_s=(1,3,3)$, followed by temporal 3D kernels $k_{t}=(3,1,1)$. The spatial attention layers are kept, and followed by temporal attention layers.
At inference, the model is given a latent noise tensor $\bm{s}_T \sim N(\bm{0}, \bm{I})^{F \times 4 \times H/8 \times W/8}$. It  produces a latent video $\bm{s}_0$ which is decoded into a pixel-space video with dimension $F \times 3 \times H \times W$ by using the \gls{vae} on each independent frame. 
We also experiment with a video \gls{vae} where we decode $\bm{s}_0$ directly to leverage the correlation between all frames of the video. We find that training a video VAE on UCF-101 directly does not yield better performance due to the small size of the training dataset, as detailed in \Cref{sec:experiments}.

The \gls{lvdm} is trained from scratch over a video dataset. Training from scratch allows us to (1) define an architecture that allows for metric comparison with other works, (2) suit our computational resources and (3) train a video model on the exact same perturbation process and noise schedule as our image model, as required for our sampling method. We detail the architectures and how we trained them in \Cref{sec:experiments}.

\subsection{Mixture of Denoising Models}
\label{sect:mix}
The major contribution of this work is to demonstrate that it is possible to combine diffusion models during the reverse diffusion process by selecting one noise prediction model at each time step. This idea allows to leverage models with different capabilities, producing better samples benefitting from all models' strengths.
To enable this, the models need to share the same diffusion training framework.
This comprises the nature of the perturbation process, which is usually the addition of Gaussian noise, as well as the same scheduling approach, to scale the perturbation at every step. 
Here, we train the models in the \gls{pndm} framework to perform $\epsilon$-prediction~\cite{ho2020denoising}. The $\epsilon$-prediction means that the neural network learns to predict the noise which has to be removed from its input.

Formally, given a real sample \(\bm{s}_0 \sim q(\bm{s}_0)\), the set of all noisy samples \(\mathcal{S}\) and its subsets of noisy samples at any given timestep \(\mathcal{S}_t\) are defined as:
\begin{align}
    \mathcal{S}_t &= \{ \bm{s}_t | \bm{s}_t = \bm{s}_0 + \bm{z}_t, \bm{z}_t \sim \mathcal{N}(\bm{0}, \sigma^2_t \bm{I}) \}, \\
    \mathcal{S} &= \{ \mathcal{S}_t | t \in \{0, \ldots, T\} \},
\label{eq:shared_space}
\end{align}
where \(\bm{z}_t\) represents the noise added to the sample \(\bm{s}_0\) at time \(t\), and \(\sigma^2_t\) is the variance of the noise at time \(t\).

Let \( \mathcal{D} = \{\mathcal{D}_1, \mathcal{D}_2, \ldots, \mathcal{D}_k\} \) be the set of denoising models.
Assuming that for all timesteps \( t \in \{0, \ldots, T\} \), any denoising model \( \mathcal{D}_i,\mathcal{D}_j \in \mathcal{D} \) is trained on the same subset \( \mathcal{S}_t \), then any model \( \mathcal{D}_{i} \) is interchangeable with \( \mathcal{D}_{j} \) in the reverse diffusion process, as they both produce outputs in the same subset \( \mathcal{S}_t \). 
In practice, this only holds if all the models are good enough to generate samples close to the true distribution of \( \mathcal{S}_t \).

In this work, we specifically focus on demonstrating that approach by combining an \gls{lidm} and an \gls{lvdm}.
As suggested by above, any group of diffusion models that share the same noise inputs and outputs distributions are compatible, regardless of their conditioning. 
By demonstrating this among models trained on vastly different tasks, \emph{i.e.}, video versus image, we empirically demonstrate that this holds even in complex cases.

While this approach works out of the box, it requires some tuning at inference.
To find out which proportion of each model we want, as well as how to best take advantage of each, we explore various parameters for our model-selection function, which we describe in \Cref{sec:experiments}.
We usually start by sampling only from the video model for a few steps, to ensure some temporal consistency, before deciding between calling one model or the other, depending on our selection function parameters.
Switching from video to image works by merging the batch and time dimensions.
This enables the denoising of all frames separately when calling the image model.
Conversely, to switch our images to a video, we undo the merging and pass the reshaped tensor into the video model to predict the noise of the entire video.

\subsection{Inference Entropy Reduction}
\label{sec:entropy}
Combining video and image models in a \gls{ddpm} setting leads to an obvious limitation when using \Cref{eq:ddpm_sampling} to sample from the models. 
When sampling from the image model, the term $\bm{z}$ adds noise to the samples with no consideration for the temporal consistency. To mitigate this, we define a new way of sampling $\bm{z}$;
\begin{align}
    \bm{z}_{\text{shared}} \sim \mathcal{N}(\bm{0}, r\bm{I}),\quad
    \bm{z}_{\text{ind}}^{f} \sim \mathcal{N}(\bm{0}, (1-r)\bm{I}), \\
    \bm{z}^{f} = \bm{z}_{\text{shared}} + \bm{z}_{\text{ind}}^{f}, \text{where}~ \bm{z}^{f} \in \mathcal{R}^{1 \times C \times H \times W},
    \label{eq:noise_red}
\end{align}
where $\bm{z}^{f}$ is the noise we add to the samples generated by the image model, $f$ is the index of the frame on the time axis, $\bm{z}_{\text{shared}}$ is the noise shared across all frames and $\bm{z}_{\text{ind}}^{f}$ is the independent noise added to each frame. $r \in [0,1]$ sets the ratio between the amount of shared and independent noise for all frames. 

Additionally, we also explore reducing the overall entropy, \emph{i.e.}, scaling down $\bm{z}$. In \Cref{eq:ddpm_sampling}, $\bm{z}$ is scaled by a factor $\sigma_t$, which defines the amount of stochasticity added back into the sample after it is denoised. We add a factor $\gamma$ to scale $\bm{z}$ such that $\sigma_t\cdot\bm{z}$ becomes $\gamma\cdot\sigma_t\cdot\bm{z}$.

We note that a similar approach has been explored in~\cite{ge2023preserve}. 
Our approach is substantially different, as we control the noise only at inference and only for the time-agnostic image model, leaving the video model untouched. 
Our \Cref{eq:noise_red} is also ratio-based and bounded, which makes it more interpretable.

\subsection{Temporal Latent Smoothing}
\label{sec:smoothing}
A major problem in current state-of-the-art diffusion models \cite{wu2022tune} is the flickering that occurs along the time axis. Some works have explored potential solutions \cite{karras2023dreampose} but we design a new approach, better suited to reduce flickering directly in the latent space. 

Our proposed approach for a latent smoothing algorithm starts by considering the latent variables of each channel separately and computes the average of each spatial position value over the time axis, which we denote $\mu_{c,h,w}$.
Following this, we compute the standard deviation of $\mu_{c,h,w}$ over $h,w$, denoted as $\sigma_c$, which acts as a measure of how much spatial variation there is in the video within each channel.
We then calculate the absolute difference between each voxel and the corresponding $\mu_{c,h,w}$ for that location, giving us a measure of how much that voxel is deviating from its average for each frame, $\Delta_{c,f,h,w}$.
We can then normalize these measures of variation using $\sigma_c$, and sum across the channels to get a single value for how much that pixel is varying within that frame in comparison to others.
A hyperparameter $c$ is then used as a threshold to determine whether the normalized variation is small enough to be close to the mean value of that location.
Latent variables with a variation measure below this threshold are therefore replaced with the mean latent code $\mu_{c,h,w}$, smoothing areas that are unlikely to belong to the action in the video.

\section{Experiments}
\label{sec:experiments}

In this section, we describe in detail how we trained all the models, starting with the diffusion models and the video autoencoder. We evaluate our ideas on the UCF-101 dataset and show the feasibility of our approach, as well as the impact of each of our post-processing steps on the final video sample quality.

\subsection{Model training}

\textbf{Dataset} We train all our models on the UCF-101 dataset~\cite{soomro2012ucf101}. The dataset consists of 13,320 videos of humans performing 101 different actions in the wild. All the videos have a resolution of $320 \times 240$ pixels and a frame rate of 25 \gls{fps} for the majority. Each video belongs to one of the 101 classes. 
For our experiments, we resample the videos to 3 resolutions, $64 \times 64$, $128 \times 128$ and $256 \times 256$. We apply $240 \times 240$ center-cropping on all the frames before resizing them to the desired resolution. We also fix the frame rate to 10 \gls{fps} for all videos. All our video models are trained to generate 16 frames-long videos.
We pre-compute the latent image representations at each resolution using the pre-trained \gls{vae} from Stable Diffusion 1.5\footnote{https://huggingface.co/runwayml/stable-diffusion-v1-5} to speed up training.

\textbf{Architectures} For all our diffusion models, we use the diffusers library\footnote{https://huggingface.co/docs/diffusers/index}. Our image models and video models share the same architecture hyperparameters, as described in \Cref{tab:architectures}. For the image model, we use a 2D downsampling layer as our first block in our UNet, followed by 3 cross-correlation blocks. We use the same strategy for the video model, but all the layers are 3D and use temporal-attention.
For our autoencoder we adopt the video \gls{vae} proposed by \cite{he2022latent}. However, since the latents are directly generated by our diffusion models we only train the decoder.

\begin{table}[t]
    \centering
    \begin{tabular}{lcc}
        \toprule
        Architecture                     & Image          & Video       \\
        \midrule
        \# channels                      & 256            & 256         \\
        Channel multiplier               & [1,2,3,4]      & [1,2,3,4]   \\
        \# residual blocks               & 2              & 2           \\
        \# channels in att. heads        & 64             & 64          \\
        \# attention heads               & 8              & 8           \\
        Cross-Attention           & [0,1,1,1]      & [0,1,1,1]   \\
        \bottomrule
    \end{tabular}
    \caption{Model architectures. The layers in the image model are all convolutional 2D layers and 2D cross-attention layers while the layers in the video model are all 3D and contain additional Temporal-Attention layers.}
    \label{tab:architectures}
\end{table}





\begin{table*}[pt]
    \begin{center}
    \begin{tabular}{l|ccc|ccc|ccc}
        \toprule
        Resolution          & \multicolumn{3}{c}{$64 \times 64$}                        & \multicolumn{3}{|c|}{$128 \times 128$}                 & \multicolumn{3}{c}{$256 \times 256$}   \\
        \midrule
        Method              & FID$\downarrow$   & FVD$\downarrow$   & IS$\uparrow$      & FID$\downarrow$ & FVD$\downarrow$ & IS$\uparrow$      & FID$\downarrow$ & FVD$\downarrow$ & IS$\uparrow$     \\
        \midrule
        Image AE~\cite{rombach2022high} & $57.09$ & $758.96$        & $4.46$            & $67.12$         & $246.46$            & $4.41$              & $75.11$         & $83.00$         & $4.32$ \\
        Video AE            & $63.86$           & $877.04$          & $4.24$           & $68.68$         & $298.71$            & $4.33$              & $83.49$             & $440.61$             & $4.13$ \\
        \midrule
        Video Model         & \cellcolor{lightgray}$23.94$  & \cellcolor{lightgray}$885.68$ & \cellcolor{lightgray}$8.28_{\pm0.26}$    & $32.63$       & $1057.34$         & $10.80_{\pm0.41}$    & \cellcolor{lightgray}$71.10$       & \cellcolor{lightgray}$1590.81$     & \cellcolor{lightgray}$7.84_{\pm0.17}$     \\
        Image Model         & $32.97$           & $1631.38$         & $7.56_{\pm0.09}$     & $17.09$       & $1840.10$         & $11.30_{\pm0.14}$    & $14.60$       & $2072.09$     & $14.51_{\pm0.12}$     \\
        \midrule
        V+I Model           & $33.8$            & $1105.39$         & $7.93_{\pm0.18}$     & $31.31$       & $1145.4$          & $11.16_{\pm0.30}$    & $31.02$       & $1808.23$     & $12.32_{\pm0.26}$ \\
        \quad+Red. Entropy  & $31.89$           & $1077.02$         & $7.91_{\pm0.20}$     & $31.70$       & $1065.15$         & $11.10_{\pm0.40}$    & $49.37$       & $1838.88$     & $10.71_{\pm0.71}$ \\
        \quad+Smoothing     & $31.79$           & $1044.23$         & $7.77_{\pm0.18}$     & \cellcolor{lightgray}$31.44$ & \cellcolor{lightgray}$1037.81$ & \cellcolor{lightgray}$11.25_{\pm0.42}$    & $50.95$       & $1872.71$     & $9.84_{\pm0.19}$ \\
        \bottomrule
    \end{tabular}
    \end{center}
    \caption{Ablation study of all our models. We evaluate each model independently, before progressively adding our improvements. The best performing video generation pipeline is highlighted in grey.}
    \label{tab:all_metrics}
\end{table*}

\textbf{Text conditioning}
For text conditioning, we use the CLIP~\cite{radford2021learning} tokenizer and text encoder. CLIP creates sequences of embeddings of dimension $768$, which we inject into the cross-correlation layers of our models.
The text we encode for the UCF-101 dataset is the name of the classes with little processing. All the class names describe a scene, but there is no rule as to how this is written, resulting in some actions described as, \emph{e.g.}, `ApplyEyeMakeup' and some object names as, \emph{e.g.},  `ParallelBars'. We apply a simple function to split the words based on the capital letters, resulting in sentences like `apply eye makeup' and `parallel bars' to which we randomly prepend either `A human', `A person' or `Someone'. While this technique would achieve subpar results if we were using a pretrained text-conditioned model, it does not impact our models as we train and infer over the same conditionings.

For completeness, we also explore our models in an unconditional setting. This is achieved by computing the embedding for the empty sentence with the text encoder, and using it as `no conditioning'.

\textbf{Training} For all models, we use the AdamW~\cite{loshchilov2017decoupled} optimizer, with a constant learning rate of $1e-4$ after 500 steps of initial warm up. We use a 30\% probability of dropping the prompt conditioning and a noise offset of 0.1\footnote{https://www.crosslabs.org//blog/diffusion-with-offset-noise} for the noise samples.

For the latent image models, we use a batch size of 1024, split over $8 \times$ A100 80Gb and no gradient accumulation. Each model is trained from scratch, for 100,000 optimization steps, resulting in training times of roughly 70, 85 and 185 GPU-hours, for the $64^2$, $128^2$ and $256^2$ models respectively.

For the latent video models, we train the models consecutively. 
We start with the $64^2$ model, trained for 100,000 steps, with a batch size of 1024, over $64 \times$ A100 80Gb. 
After the low-resolution $64^2$ model is trained, we re-use the weights and continue training on the $128^2$ resolution for an additional 50,000 steps with the exact same configuration.
Finally, we train the $256^2$ model starting from the $128^2$ model.
Due to training instability leading to \texttt{NaN}-values, we increase the learning to $2e-4$ and the $\epsilon$ value in the AdamW optimizer from $1e-8$ to $1e-04$.
For this model, we reduce the per-device batch size by 2 and set gradient accumulation to 2, in order to keep a global batch size of 1024.
The model is trained for another 50,000 steps.
This results in 2400 GPU-hours for the $64^2$ model, 1500 additional hours for $128^2$ model and another 2000 for the final 50,000 steps at resolution $256^2$.

\subsection{Models evaluation}
For each model, we perform iterative hyperparameter searches.
For the video models, we keep the hyperparameters lowering the \gls{fvd} the most, while for the image model, we focus on the \gls{fid}.
The first hyperparameter search focuses on the best combination of guidance scale and sampler.
We evaluate guidance scales [0, 1, 2, 4, 6, 8, 10] and samplers \gls{ddpm}, \gls{ddim} and \gls{pndm}. After establishing the best guidance scale and sampler, we look for the best number of sampling steps, evaluating 20, 30, 40, 50, 60, 75, 100 and 200 steps. We find that, for all models, the combination of the \gls{ddpm} scheduler, guidance scale 2.0 and 50 sampling steps either yields the best results or falls within the margin of error. We therefore use these parameters for all the models. 
We then explore the model selection for the mixture of denoising models, which we describe in more detail in \Cref{sec:exp_mixture}. Once we have defined the best way to select the models, we move on to the post-processing.
First, we explore whether reducing the entropy in the reverse diffusion process (\Cref{eq:ddpm_sampling}) helps to better maintain the temporal consistency. We check the metrics for parameter $\gamma$ (\Cref{sec:entropy}), as well as the ratio $r$ for both the video model ($r_v$) and the image model ($r_i$). We test $\gamma \in [0.5, 0.2, 0.1, 0.05, 0.02, 0.01]$ and $r_v,r_i \in [0.2, 0.4, 0.6, 0.8, 1.0]$.
We discover that using correlated noise does not help with metrics, thus $r_v$ and $r_i$ are set to 0 for all models.
For $\gamma$, we find that the $64^2$ model works better with $\gamma=0.02$, the $128^2$ model with $\gamma=0.1$ and the $256^2$ model with $\gamma=1.0$.
For the smoothing post-processing, we explore three algorithms: Gaussian smoothing, uniform smoothing, and the custom-designed approach presented in \Cref{sec:smoothing}.
We find that our smoothing strategy works the best, with a smoothing factor of 2 for all models.
All experiments are seeded, to enable reproducibility and a fair comparison.

\begin{figure}
    \centering
    \begin{picture}(100,100)
    \put(-60,0){
    \includegraphics[width=0.45\textwidth]{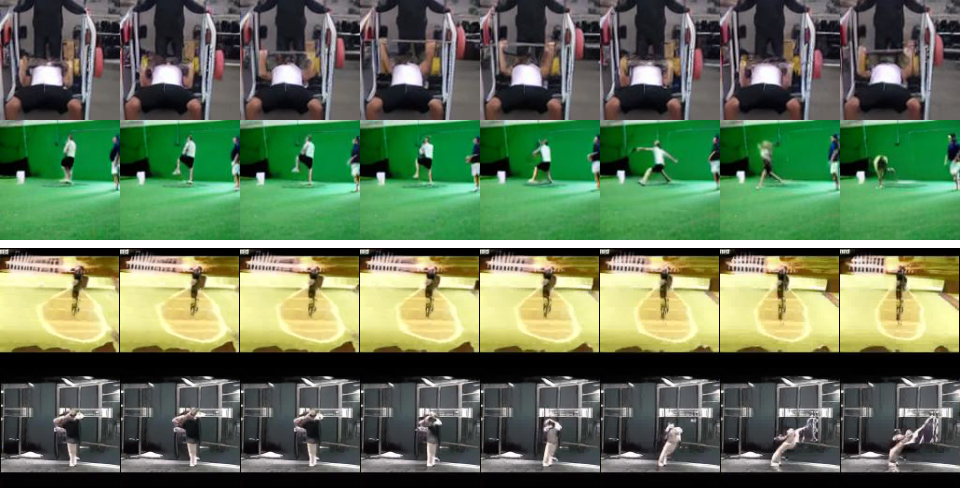}}
    \put(-70,65){\rotatebox{90}{{\footnotesize JVID (ours)}}}
    \put(-70,5){\rotatebox{90}{{\footnotesize StyleGAN-V}}}
    \end{picture}
    \caption{Our JVID model (top two rows) demonstrates higher image quality and temporal consistency compared to StyleGAN-V~\cite{skorokhodov2021stylegan} (bottom two rows).}
    \label{fig:comparison}
\end{figure}

Regarding metric computation, we follow the StyleGAN-V protocol~\cite{skorokhodov2021stylegan}.
We compute the \gls{fid} over 50,000 images, where the images are part of 16-frames long videos, the \gls{fvd}$_{16}$ over 2048 videos and the \gls{is} over $10 \times 5,000$ images, and report the mean and standard deviation.
We note that StyleGAN-V~\cite{skorokhodov2021stylegan} re-evaluated some of the models we compare against. 
To make sure to provide a fair comparison, we compare our results to both the metrics claimed by each individual work, as well as the metrics recomputed by StyleGAN-V. We report all metrics in \Cref{tab:stylegan}, \Cref{tab:ucf_cond} and \Cref{tab:ucf_unc}, and show qualitative results in \Cref{fig:samples} and \Cref{fig:comparison}.

\begin{table}[pt]
    \centering
    \begin{tabular}{lc}
        \toprule
        Method & \gls{fvd}$_{16}$ \\
        \midrule
        MoCoGAN \cite{tulyakov2018mocogan}                      & 2886.9 \\
        + StyleGAN2 backbone \cite{skorokhodov2021stylegan}     & 1821.4 \\
        MoCoGAN-HD \cite{tian2021good}                          & 1729.6 \\
        VideoGPT \cite{yan2021videogpt}†                        & 2880.6 \\
        DIGAN \cite{yu2022generating}                           & 1630.2 \\
        StyleGAN-V \cite{skorokhodov2021stylegan}               & 1431.0 \\
        JVID (ours)                                             & 1590.81\\
        \bottomrule
    \end{tabular}
    \caption{We compare our method to StyleGAN-V~\cite{skorokhodov2021stylegan} recomputed metrics following the StyleGAN-V protocol on resolution $256^2$.}
    \label{tab:stylegan}
\end{table}

\begin{table}[pt]
    \centering 
    \begin{tabular}{lccc}
        \toprule
        Method                                  & Res.& \gls{fvd} ($\downarrow$) & \gls{is} ($\uparrow$)   \\
        \midrule
        TGANv2 \cite{saito2020train}            & 128 & -              & $26.60_{\pm 0.47}$  \\
        DIGAN \cite{yu2022generating}           & 128 & $577_{\pm 22}$ & $32.70$             \\
        MoCoGAN-HD \cite{tian2021good}          & 256 & $700_{\pm 24}$ & $33.95_{\pm 0.25}$  \\
        CogVideo \cite{hong2022cogvideo}        & 160 & $626$          & $50.46$             \\
        VDM \cite{ho2022video}                  & 64  & -              & $57.80_{\pm 1.3}$   \\
        TATS-base \cite{ge2022long}             & 128 & $278_{\pm 11}$ & $79.28_{\pm 0.38}$  \\
        Make-A-Video \cite{singer2022make}      & 256 & $81.25$        & $82.55$             \\
        \midrule
        JVID (ours)                             & 64  & $885.68$       & $8.28_{\pm0.26}$    \\
        JVID (ours)                             & 128 & $1037.81$      & $11.25_{\pm0.42}$   \\
        JVID (ours)                             & 256 & $1590.81$            & $7.84_{\pm0.17}$             \\
        \bottomrule
    \end{tabular}
    \caption{Text-conditioned video generation on UCF-101. All models are trained or fine-tuned on UCF-101.}
    \label{tab:ucf_cond}
\end{table}

\begin{table}[pt]
    \centering 
    \begin{tabular}{lccc}
        \toprule
        Method                                      & Res.              & \gls{fvd} (\(\downarrow\))  & \gls{is} (\(\uparrow\))     \\
        \midrule
        TGAN \cite{saito2017temporal}               & $128$             & -                     & $15.83_{\pm.18}$       \\
        LDVD-GAN \cite{kahembwe2020lower}           & $64$            & -                     & $22.91_{\pm.19}$       \\
        VideoGPT \cite{yan2021videogpt}             & $128$             & -                     & $24.69_{\pm.30}$       \\
        MoCoGAN-HD \cite{tulyakov2018mocogan}       & $256$             & $838$                 & $32.36$                \\
        DIGAN \cite{yu2022generating}               & $128$             & $655_{\pm22}$         & $29.71_{\pm.53}$       \\
        CCVS \cite{le_moing2021ccvs}                & $128$             & $386_{\pm15}$         & $24.47_{\pm.13}$       \\
        StyleGAN-V \cite{skorokhodov2021stylegan}   & $256$             & -                     & $23.94_{\pm.73}$       \\
        VDM \cite{ho2022video}                      & $64$              & -                     & $57.00_{\pm.62}$       \\
        TATS \cite{ge2022long}                      & $128$             & $430_{\pm18}$         & $57.63_{\pm.73}$       \\
        PYoCo \cite{ge2023preserve}                 & $64$            & $310_{\pm13}$         & $60.01_{\pm.51}$       \\
        \midrule
        JVID (ours)                                 & $64$              & $2143.16$             & $6.02_{\pm0.15}$       \\
        JVID (ours)                                 & $128$             & $1257.07$             & $10.25_{\pm0.39}$      \\
        JVID (ours)                                 & $256$             & $2681.54$             & $7.16_{\pm0.19}$      \\
        \bottomrule
    \end{tabular}
    \caption{Unconditional video generation on UCF-101.}
    \label{tab:ucf_unc}
\end{table}

\subsection{Mixture of Denoising Models}
\label{sec:exp_mixture}
\begin{figure}[pt]
    \centering
    \includegraphics{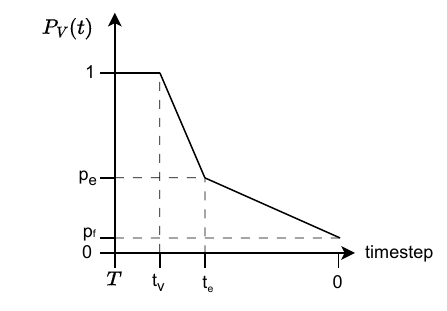}
    \caption{Model selection probability function. Parameters $t_v$, $t_e$, $p_e$, $p_f$ are determined empirically.}
    \label{fig:sampling}
\end{figure}

The main contribution of this work is the demonstration that different diffusion models, trained on identical input and output distributions, are compatible during the reverse diffusion process. 
Here we present the reasoning behind our mixing strategy. It decides which model to call at which step, in the reverse diffusion process. 
When mixing a time-aware video and a time-agnostic image model, we consider that the video model will take a noisy input and denoise it with the objective of enforcing temporal consistency, while the image model will denoise the same noisy input and push it towards better image quality. 
When denoising with our image model, the model is deterministic, but ignores the temporal consistency, resulting in samples that slowly degrade the images' coherence over the time axis. Thus, the idea is to sample from both models at optimal timesteps to maintain both temporal consistency and image quality.
From early experimentation, we notice that starting by sampling exclusively from the video model, is central to setting up a latent representation with a good temporal consistency.
After that, we randomly select the model which we sample from, given a probability $P_V(t)$ to sample from the video model.
It appears that abruptly altering the model sampling probability causes a degradation in video quality. Therefore, we opt for a gradual linear interpolation between the initial and final values of $P_V(t)$.
It also turns out that sampling only from the image model for a given amount of final steps led to noticeable damage to the temporal consistency, and as such some probability of sampling from the video model is kept for the entire sampling duration.
We opt for a piece-wise affine function, to allow for different probability gradients, as shown in Figure \ref{fig:sampling}. 
This approach was selected to enhance flexibility during the sampling process, enabling the image model to exert a greater influence in the initial stages without prematurely diminishing the probability of the video model's contribution. This concept can be formally expressed as 
\[
P_V(t) = \begin{cases} 
1.0, & \text{if } t \geq t_v \\
1 + \left( \frac{t - t_v}{t_e - t_v} \right) \cdot (p_e - 1), & \text{if } t_v > t \geq t_e \\
p_e + \left( \frac{t - t_e}{1 - t_e} \right) \cdot (p_f - p_e), & \text{otherwise.}
\end{cases}
\]

We report the best model section hyperparameters for each resolution in \Cref{tab:modsel}.

\begin{table}[t]
    \centering
    \begin{tabular}{ccccc}
        \toprule
        Resolution      & $t_v$ & $t_e$ & $p_e$ & $p_f$ \\
        \midrule
        $64 \times 64$  & $0.2$ & $0.7$ & $0.3$ & $0.3$ \\
        $128 \times 128$& $0.4$ & $0.7$ & $0.4$ & $0.1$ \\
        $256 \times 256$& $0.1$ & $0.6$ & $0.2$ & $0.1$ \\
        \bottomrule
    \end{tabular}
    \caption{Best model selection hyperparameters.}
    \label{tab:modsel}
\end{table}

\section{Discussion}


We acknowledge that the models used in this work do not beat state-of-the-art approaches. This is due to the considerable amount of compute necessary to match these models' performance. For example, Make-A-Video~\cite{singer2022make} reaches an \gls{fvd} of 81.25. However, their model is pre-trained over millions of samples before being fine-tuned on UCF-101, which is not something usually achievable with academic resources. Additionally, the very few papers which do release their models and code were not immediately compatible with our method, requiring us to fully retrain all our models. 

When fairly comparing our work, especially with the results from \Cref{tab:stylegan}, we come close to the state-of-the-art when training on similar amounts of data.
This leads us to believe that our mixture of denoising models should be explored further in more settings, and at a greater scale.

\noindent\textbf{Environmental Impact. } 
Our work involved substantial energy consumption, utilizing 8,000 hours of A40 energy and 24,000 hours of A100 energy, cumulatively amounting to a significant 12,000 kilowatt-hours of electricity used.
A Tesla Model 3 could be powered for 55,800 miles with this amount of energy. 
This level of energy usage underscores the importance of considering the environmental impact of such projects, particularly in terms of their carbon footprint and contribution to energy consumption.
Therefore, we will publish our model weights together with all the code so that any subsequent work can continue from this point. 

\section{Conclusion}
\label{sec:discussion}

This work presents a new way to combine diffusion models during inference.
Our results show that we can indeed generate samples while using two independently trained diffusion models. Additionally, our approach has minimal requirements and should synergize with many existing conditioning approaches, which is what we intend to explore in future works.



{
    \small
    \bibliographystyle{ieeenat_fullname}
    \bibliography{main}
}


\end{document}